\documentclass{article}

     \usepackage[preprint]{neurips_2024}

\usepackage[utf8]{inputenc} 
\usepackage[T1]{fontenc}    
\usepackage[hyphens]{url}
\usepackage{hyperref}
\usepackage[hyphenbreaks]{breakurl}
\usepackage{booktabs}       
\usepackage{amsfonts}       
\usepackage{nicefrac}       
\usepackage{microtype}      
\usepackage{xcolor}         
\usepackage{longtable}         
\usepackage{tikz}
\usepackage{natbib}
\usepackage{amsmath}
\usepackage{amssymb}
\usepackage{amsthm}
\usepackage{algorithm}
\usepackage{algorithmic}
\usepackage{float}

\setcitestyle{numbers}

\title{Compliance Cards: Automated EU AI Act Compliance Analyses amidst a Complex AI Supply Chain}

\author{%
  Bill Marino \textsuperscript{1,2}\thanks{Corresponding Author: \texttt{wlm27@cam.ac.uk}}, Yaqub Chaudhary \textsuperscript{2}, Yulu Pi \textsuperscript{2,4}, Rui-Jie Yew \textsuperscript{5}\\
  \textbf{Preslav Aleksandrov \textsuperscript{1}},\\
  \textbf{Carwyn Rahman \textsuperscript{3}, William F. Shen \textsuperscript{1}}\\
  \textbf{Isaac Robinson \textsuperscript{6}, Nicholas D. Lane \textsuperscript{1}} \\
  \\
  \textsuperscript{1} University of Cambridge, UK \hspace*{10pt}
  \textsuperscript{2} Leverhulme Center for the Future of Intelligence, UK\\
  \textsuperscript{3} Protocol Consulting, Ltd., UK \hspace*{10pt}
  \textsuperscript{4} University of Warwick, UK\\
  \textsuperscript{5} Brown University, US  \hspace*{10pt}\\
  \textsuperscript{6} Oxford University, UK  
}

\begin{document}

\maketitle

\begin{abstract}
  As the AI supply chain grows more complex, AI systems and models are increasingly likely to incorporate multiple internally- or externally-sourced components such as datasets and (pre-trained) models. In such cases, determining whether or not the aggregate AI system or model complies with the EU AI Act (``AIA'') requires a multi-step process in which compliance-related information about both the AI system or model \emph{and} all its component parts is: (1) gathered, potentially from multiple arms-length sources; (2) harmonized, if necessary; (3) inputted into an analysis that looks across all of it to render a compliance prediction. Because this process is so complex and time-consuming, it threatens to overburden the limited compliance resources of the AI providers (i.e., developers) who bear much of the responsibility for complying with the AIA. It also renders rapid or real-time compliance analyses infeasible in many AI development scenarios where they would be beneficial to providers. To address these shortcomings, we introduce a complete system for automating provider-side AIA compliance analyses amidst a complex AI supply chain. This system has two key elements. First is an interlocking set of computational, multi-stakeholder transparency artifacts that capture AIA-specific metadata about both: (1) the provider's overall AI system or model; and (2) the datasets and pre-trained models it incorporates as components. Second is an algorithm that operates across all those artifacts to render a real-time prediction about whether or not the aggregate AI system or model complies with the AIA. All told, this system promises to dramatically facilitate and democratize provider-side AIA compliance analyses (and, perhaps by extension, provider-side AIA compliance).
\end{abstract}


\section{Introduction}
The EU AI Act (``AIA''), which recently entered into force \citep{techcrunchEUsGets}, sets forth harmonized requirements for AI systems and models (which we collectively call ``AI projects'') \citep[Art. 1(2)]{europa}.

Importantly, these requirements not only regulate aspects of the AI projects themselves, but also aspects of the datasets and (pre-trained) models that those AI projects incorporate as components. The practical impact of this, amidst an AI landscape where both AI projects \citep{berkeleyShiftFrom} and the supply chain behind them \citep{adalovelaceinstituteAllocatingAccountability, ceps} are growing more complex, is that determining whether or not an AI project complies with the AIA often requires the following multi-step procedure (Figure~\ref{fig:statusquo}): 

\begin{enumerate}
\item{Gathering information about the AI project \textit{and all its component datasets and models}, potentially from multiple internal or external (and sometimes arms-length) sources;}
\item{Optionally harmonizing that information (if its sources used different formats to capture it);}
\item{Waging an analysis that looks across all of this collected/harmonized information in order to render a prediction about the compliance level of the aggregate AI project.}
\end{enumerate}

\begin{figure}[ht]

\begin{center}

\tikzset{every picture/.style={line width=0.75pt}} 

\begin{tikzpicture}[x=0.75pt,y=0.75pt,yscale=-1,xscale=1]

\draw  [fill={rgb, 255:red, 255; green, 255; blue, 255 }  ,fill opacity=1 ] (78,96.8) .. controls (78,83.93) and (88.43,73.5) .. (101.3,73.5) -- (171.2,73.5) .. controls (184.07,73.5) and (194.5,83.93) .. (194.5,96.8) -- (194.5,236.7) .. controls (194.5,249.57) and (184.07,260) .. (171.2,260) -- (101.3,260) .. controls (88.43,260) and (78,249.57) .. (78,236.7) -- cycle ;
\draw  [fill={rgb, 255:red, 248; green, 231; blue, 28 }  ,fill opacity=1 ] (171,213.65) -- (171,221.35) .. controls (171,222.26) and (167.87,223) .. (164,223) .. controls (160.13,223) and (157,222.26) .. (157,221.35) -- (157,213.65) .. controls (157,212.74) and (160.13,212) .. (164,212) .. controls (167.87,212) and (171,212.74) .. (171,213.65) .. controls (171,214.56) and (167.87,215.3) .. (164,215.3) .. controls (160.13,215.3) and (157,214.56) .. (157,213.65) ;
\draw  [fill={rgb, 255:red, 248; green, 231; blue, 28 }  ,fill opacity=1 ] (181,227.65) -- (181,235.35) .. controls (181,236.26) and (177.87,237) .. (174,237) .. controls (170.13,237) and (167,236.26) .. (167,235.35) -- (167,227.65) .. controls (167,226.74) and (170.13,226) .. (174,226) .. controls (177.87,226) and (181,226.74) .. (181,227.65) .. controls (181,228.56) and (177.87,229.3) .. (174,229.3) .. controls (170.13,229.3) and (167,228.56) .. (167,227.65) ;
\draw  [fill={rgb, 255:red, 248; green, 231; blue, 28 }  ,fill opacity=1 ] (161,227.65) -- (161,235.35) .. controls (161,236.26) and (157.87,237) .. (154,237) .. controls (150.13,237) and (147,236.26) .. (147,235.35) -- (147,227.65) .. controls (147,226.74) and (150.13,226) .. (154,226) .. controls (157.87,226) and (161,226.74) .. (161,227.65) .. controls (161,228.56) and (157.87,229.3) .. (154,229.3) .. controls (150.13,229.3) and (147,228.56) .. (147,227.65) ;

\draw  [fill={rgb, 255:red, 255; green, 255; blue, 255 }  ,fill opacity=1 ] (220.5,67.5) -- (285,67.5) -- (285,132) -- (220.5,132) -- cycle ;
\draw  [fill={rgb, 255:red, 192; green, 221; blue, 251 }  ,fill opacity=1 ] (220.5,67.5) -- (285,67.5) -- (285,90) -- (220.5,90) -- cycle ;
\draw  [fill={rgb, 255:red, 80; green, 227; blue, 194 }  ,fill opacity=1 ] (90.34,215.72) -- (93.46,212.6) -- (104,212.6) -- (104,219.88) -- (100.88,223) -- (90.34,223) -- cycle ; \draw   (104,212.6) -- (100.88,215.72) -- (90.34,215.72) ; \draw   (100.88,215.72) -- (100.88,223) ;
\draw  [fill={rgb, 255:red, 80; green, 227; blue, 194 }  ,fill opacity=1 ] (99.34,231.72) -- (102.46,228.6) -- (113,228.6) -- (113,235.88) -- (109.88,239) -- (99.34,239) -- cycle ; \draw   (113,228.6) -- (109.88,231.72) -- (99.34,231.72) ; \draw   (109.88,231.72) -- (109.88,239) ;
\draw  [fill={rgb, 255:red, 80; green, 227; blue, 194 }  ,fill opacity=1 ] (110.34,215.72) -- (113.46,212.6) -- (124,212.6) -- (124,219.88) -- (120.88,223) -- (110.34,223) -- cycle ; \draw   (124,212.6) -- (120.88,215.72) -- (110.34,215.72) ; \draw   (120.88,215.72) -- (120.88,223) ;

\draw  [fill={rgb, 255:red, 229; green, 206; blue, 249 }  ,fill opacity=1 ] (418,83) -- (477.5,83) -- (477.5,255) -- (418,255) -- cycle ;

\draw    (194.5,102.5) -- (213.2,102.5) ;
\draw [shift={(215.2,102.5)}, rotate = 180] [color={rgb, 255:red, 0; green, 0; blue, 0 }  ][line width=0.75]    (10.93,-3.29) .. controls (6.95,-1.4) and (3.31,-0.3) .. (0,0) .. controls (3.31,0.3) and (6.95,1.4) .. (10.93,3.29)   ;
\draw    (194.8,231.5) -- (213.5,231.5) ;
\draw [shift={(215.5,231.5)}, rotate = 180] [color={rgb, 255:red, 0; green, 0; blue, 0 }  ][line width=0.75]    (10.93,-3.29) .. controls (6.95,-1.4) and (3.31,-0.3) .. (0,0) .. controls (3.31,0.3) and (6.95,1.4) .. (10.93,3.29)   ;
\draw  [fill={rgb, 255:red, 255; green, 255; blue, 255 }  ,fill opacity=1 ] (235.5,207.5) -- (300,207.5) -- (300,272) -- (235.5,272) -- cycle ;
\draw  [fill={rgb, 255:red, 192; green, 221; blue, 251 }  ,fill opacity=1 ] (235.5,207.5) -- (300,207.5) -- (300,230) -- (235.5,230) -- cycle ;
\draw  [fill={rgb, 255:red, 255; green, 255; blue, 255 }  ,fill opacity=1 ] (220.5,146.5) -- (285,146.5) -- (285,211) -- (220.5,211) -- cycle ;
\draw  [fill={rgb, 255:red, 192; green, 221; blue, 251 }  ,fill opacity=1 ] (221,146.5) -- (285,146.5) -- (285,169) -- (221,169) -- cycle ;
\draw    (303.5,102.5) -- (322.2,102.5) ;
\draw [shift={(324.2,102.5)}, rotate = 180] [color={rgb, 255:red, 0; green, 0; blue, 0 }  ][line width=0.75]    (10.93,-3.29) .. controls (6.95,-1.4) and (3.31,-0.3) .. (0,0) .. controls (3.31,0.3) and (6.95,1.4) .. (10.93,3.29)   ;
\draw    (303.8,231.5) -- (322.5,231.5) ;
\draw [shift={(324.5,231.5)}, rotate = 180] [color={rgb, 255:red, 0; green, 0; blue, 0 }  ][line width=0.75]    (10.93,-3.29) .. controls (6.95,-1.4) and (3.31,-0.3) .. (0,0) .. controls (3.31,0.3) and (6.95,1.4) .. (10.93,3.29)   ;
\draw    (481.8,166.5) -- (500.5,166.5) ;
\draw [shift={(502.5,166.5)}, rotate = 180] [color={rgb, 255:red, 0; green, 0; blue, 0 }  ][line width=0.75]    (10.93,-3.29) .. controls (6.95,-1.4) and (3.31,-0.3) .. (0,0) .. controls (3.31,0.3) and (6.95,1.4) .. (10.93,3.29)   ;
\draw  [fill={rgb, 255:red, 217; green, 239; blue, 195 }  ,fill opacity=1 ] (332,83) -- (388.5,83) -- (388.5,255) -- (332,255) -- cycle ;
\draw    (391.5,165.5) -- (410.2,165.5) ;
\draw [shift={(412.2,165.5)}, rotate = 180] [color={rgb, 255:red, 0; green, 0; blue, 0 }  ][line width=0.75]    (10.93,-3.29) .. controls (6.95,-1.4) and (3.31,-0.3) .. (0,0) .. controls (3.31,0.3) and (6.95,1.4) .. (10.93,3.29)   ;

\draw (100,92) node [anchor=north west][inner sep=0.75pt]   [align=left] {\begin{minipage}[lt]{50.34pt}\setlength\topsep{0pt}
\begin{center}
\textbf{AI Project}
\end{center}

\end{minipage}};
\draw (506,142) node [anchor=north west][inner sep=0.75pt]   [align=left] {\begin{minipage}[lt]{52.71pt}\setlength\topsep{0pt}
\begin{center}
\textbf{{\scriptsize Compliance}}\\\textbf{{\scriptsize Determination }}\\\textbf{{\scriptsize for AI Project}}
\end{center}

\end{minipage}};
\draw (229,145.5) node [anchor=north west][inner sep=0.75pt]   [align=left] {\begin{minipage}[lt]{30.77pt}\setlength\topsep{0pt}
\begin{center}
\textbf{{\tiny Information}}
\end{center}

\end{minipage}};
\draw (231,173) node [anchor=north west][inner sep=0.75pt]  [font=\normalsize] [align=left] {\begin{minipage}[lt]{31.05pt}\setlength\topsep{0pt}
\begin{center}
{\tiny \textbf{Component}}\\{\tiny \textbf{Models}}
\end{center}

\end{minipage}};
\draw (245,206.5) node [anchor=north west][inner sep=0.75pt]   [align=left] {\begin{minipage}[lt]{30.77pt}\setlength\topsep{0pt}
\begin{center}
\textbf{{\tiny Information}}
\end{center}

\end{minipage}};
\draw (243,233) node [anchor=north west][inner sep=0.75pt]  [font=\normalsize] [align=left] {\begin{minipage}[lt]{32.47pt}\setlength\topsep{0pt}
\begin{center}
{\tiny \textbf{Component }}\\{\tiny \textbf{Datasets}}
\end{center}

\end{minipage}};
\draw (234,99.5) node [anchor=north west][inner sep=0.75pt]   [align=left] {\begin{minipage}[lt]{27.95pt}\setlength\topsep{0pt}
\begin{center}
\textbf{{\tiny AI Project}}
\end{center}

\end{minipage}};
\draw (229,66.5) node [anchor=north west][inner sep=0.75pt]   [align=left] {\begin{minipage}[lt]{30.77pt}\setlength\topsep{0pt}
\begin{center}
\textbf{{\tiny Information}}
\end{center}

\end{minipage}};
\draw (425,32) node [anchor=north west][inner sep=0.75pt]   [align=left] {\textbf{{\scriptsize Step 3}}};
\draw (336,146) node [anchor=north west][inner sep=0.75pt]  [font=\normalsize] [align=left] {\begin{minipage}[lt]{30.77pt}\setlength\topsep{0pt}
\begin{center}
{\tiny \textbf{Harmonize }}\\{\tiny \textbf{information}}
\end{center}

\end{minipage}};
\draw (230,32) node [anchor=north west][inner sep=0.75pt]   [align=left] {\textbf{{\scriptsize Step 1}}};
\draw (328,32) node [anchor=north west][inner sep=0.75pt]   [align=left] {\begin{minipage}[lt]{34.4pt}\setlength\topsep{0pt}
\begin{center}
\textbf{{\scriptsize Step 2 }}\\{\scriptsize (optional{\fontfamily{pcr}\selectfont )}}
\end{center}

\end{minipage}};
\draw (206,276) node [anchor=north west][inner sep=0.75pt]   [align=left] {{\tiny May come from different sources}};
\draw (423,132) node [anchor=north west][inner sep=0.75pt]  [font=\normalsize] [align=left] {\begin{minipage}[lt]{33.04pt}\setlength\topsep{0pt}
\begin{center}
{\tiny \textbf{Compliance }}\\{\tiny \textbf{Analysis }}\\{\tiny \textbf{across all }}\\{\tiny \textbf{information}}
\end{center}

\end{minipage}};
\draw (86,173) node [anchor=north west][inner sep=0.75pt]  [font=\normalsize] [align=left] {\begin{minipage}[lt]{28.79pt}\setlength\topsep{0pt}
\begin{center}
{\tiny \textbf{Integrated }}\\{\tiny \textbf{Models}}
\end{center}

\end{minipage}};
\draw (143,173) node [anchor=north west][inner sep=0.75pt]  [font=\normalsize] [align=left] {\begin{minipage}[lt]{28.79pt}\setlength\topsep{0pt}
\begin{center}
{\tiny \textbf{Integrated }}\\{\tiny \textbf{Datasets}}
\end{center}

\end{minipage}};

\end{tikzpicture}

\end{center}
\caption{\textit{\textbf{Today's AIA Compliance Analysis Procedure}: In Step 1, information about the overall AI Project as well as its component Models and Datasets is gathered from internal and/or external sources. In Step 2, this information is optionally harmonized. In Step 3, an analysis looks across all of the gathered/harmonized information to render a compliance analysis for the overall AI Project.}}\label{fig:statusquo}
\end{figure}

Needless to say, this process can be very time-consuming \footnote{For example, the Impact Assessment accompanying the AIA estimated that the process of ``review[ing] documentation'' to verify the compliance of AI systems that are safety components of products would take ``between one and two and a half days'' and cost 3000€ – 7500€ \citep[68-69]{europa2}.}. It therefore threatens to strain the limited compliance resources of the AI providers (i.e., developers) \footnote{The AIA defines several AI ``operator'' roles, with different obligations for each \citep[Art. 3(8)]{europa}. Among these, the ``provider'', which the AIA defines as anyone that develops an AI project (or that has an AI project developed) and places it on the market or puts it into service under their own name or trademark \citep[Art. 3(3)]{europa}, is awarded the highest level of responsibility for satisfying the AIA's requirements \citep[Arts. 2(1), 3(3)]{europa, edri}.} who ``bear the largest share of obligations under the [AIA]'' \citep{cdtPublicAuthorities}. Importantly, it also renders real-time compliance assessments impossible in the many scenarios (listed in Section~\ref{sec:workflows}) where they would be highly beneficial to these providers.

To help solve these problems, we introduce the Compliance Cards system for automated provider-side AIA compliance analyses. It is built around two innovative technical elements: 

\begin{itemize}
\item{\textbf{Compliance Cards}: an  interlocking set of open-source transparency artifacts (akin to Model or Data Cards \citep{DBLP:conf/fat/MitchellWZBVHSR19, 10.1145/3531146.3533231}) that capture, \emph{in a computational format conductive to algorithmic manipulation}, compliance-related metadata about both: (1) a provider's AI project; and (2) any individual datasets and models it integrates;} 
\item{\textbf{Compliance Cards Algorithm}: an algorithm that manipulates the metadata in a set of Compliance Cards to render a run-time prediction about whether the provider's AI project complies with the AIA.}
\end{itemize}

The vision is that the Compliance Cards artifacts for a given AI project (and its component datasets and models) will be populated, in advance and asynchronously, by the parties best-equipped to do so. Then, the artifacts can be assembled and inputted into the Compliance Cards Algorithm by anyone who wishes to analyze whether the provider's AI project satisfies the AIA's requirements. Because the artifacts are pre-harmonized, our process eliminates the harmonization step in (Figure~\ref{fig:statusquo}). Because they can be operated on by the accompanying algorithm, we also automate the analysis step in  (Figure~\ref{fig:statusquo}).\footnote{The AIA itself champions the notion of an automated compliance analyses, albeit in relation to another type of AI operator: ``The AI Office shall develop a template for a questionnaire, including through an automated tool, to facilitate deployers in complying with their obligations...in a simplified manner'' \citep[Art. 27(5)]{europa}.} The net result is the streamlined AIA compliance analysis process depicted in (Figure~\ref{fig:ccanalysis}), which promises to accelerate and democratize compliance assessments for providers, their third party assessors, regulators, or anyone else (see Section~\ref{sec:who}).

\begin{figure}[ht]

\begin{center}

\tikzset{every picture/.style={line width=0.75pt}} 

\begin{tikzpicture}[x=0.75pt,y=0.75pt,yscale=-1,xscale=1]

\draw  [fill={rgb, 255:red, 255; green, 255; blue, 255 }  ,fill opacity=1 ] (78,96.8) .. controls (78,83.93) and (88.43,73.5) .. (101.3,73.5) -- (171.2,73.5) .. controls (184.07,73.5) and (194.5,83.93) .. (194.5,96.8) -- (194.5,236.7) .. controls (194.5,249.57) and (184.07,260) .. (171.2,260) -- (101.3,260) .. controls (88.43,260) and (78,249.57) .. (78,236.7) -- cycle ;
\draw  [fill={rgb, 255:red, 248; green, 231; blue, 28 }  ,fill opacity=1 ] (171,213.65) -- (171,221.35) .. controls (171,222.26) and (167.87,223) .. (164,223) .. controls (160.13,223) and (157,222.26) .. (157,221.35) -- (157,213.65) .. controls (157,212.74) and (160.13,212) .. (164,212) .. controls (167.87,212) and (171,212.74) .. (171,213.65) .. controls (171,214.56) and (167.87,215.3) .. (164,215.3) .. controls (160.13,215.3) and (157,214.56) .. (157,213.65) ;
\draw  [fill={rgb, 255:red, 248; green, 231; blue, 28 }  ,fill opacity=1 ] (181,227.65) -- (181,235.35) .. controls (181,236.26) and (177.87,237) .. (174,237) .. controls (170.13,237) and (167,236.26) .. (167,235.35) -- (167,227.65) .. controls (167,226.74) and (170.13,226) .. (174,226) .. controls (177.87,226) and (181,226.74) .. (181,227.65) .. controls (181,228.56) and (177.87,229.3) .. (174,229.3) .. controls (170.13,229.3) and (167,228.56) .. (167,227.65) ;
\draw  [fill={rgb, 255:red, 248; green, 231; blue, 28 }  ,fill opacity=1 ] (161,227.65) -- (161,235.35) .. controls (161,236.26) and (157.87,237) .. (154,237) .. controls (150.13,237) and (147,236.26) .. (147,235.35) -- (147,227.65) .. controls (147,226.74) and (150.13,226) .. (154,226) .. controls (157.87,226) and (161,226.74) .. (161,227.65) .. controls (161,228.56) and (157.87,229.3) .. (154,229.3) .. controls (150.13,229.3) and (147,228.56) .. (147,227.65) ;

\draw  [fill={rgb, 255:red, 80; green, 227; blue, 194 }  ,fill opacity=1 ] (90.34,215.72) -- (93.46,212.6) -- (104,212.6) -- (104,219.88) -- (100.88,223) -- (90.34,223) -- cycle ; \draw   (104,212.6) -- (100.88,215.72) -- (90.34,215.72) ; \draw   (100.88,215.72) -- (100.88,223) ;
\draw  [fill={rgb, 255:red, 80; green, 227; blue, 194 }  ,fill opacity=1 ] (99.34,231.72) -- (102.46,228.6) -- (113,228.6) -- (113,235.88) -- (109.88,239) -- (99.34,239) -- cycle ; \draw   (113,228.6) -- (109.88,231.72) -- (99.34,231.72) ; \draw   (109.88,231.72) -- (109.88,239) ;
\draw  [fill={rgb, 255:red, 80; green, 227; blue, 194 }  ,fill opacity=1 ] (110.34,215.72) -- (113.46,212.6) -- (124,212.6) -- (124,219.88) -- (120.88,223) -- (110.34,223) -- cycle ; \draw   (124,212.6) -- (120.88,215.72) -- (110.34,215.72) ; \draw   (120.88,215.72) -- (120.88,223) ;

\draw  [fill={rgb, 255:red, 229; green, 206; blue, 249 }  ,fill opacity=1 ] (377,79) -- (436.5,79) -- (436.5,251) -- (377,251) -- cycle ;
\draw    (194.5,102.5) -- (213.2,102.5) ;
\draw [shift={(215.2,102.5)}, rotate = 180] [color={rgb, 255:red, 0; green, 0; blue, 0 }  ][line width=0.75]    (10.93,-3.29) .. controls (6.95,-1.4) and (3.31,-0.3) .. (0,0) .. controls (3.31,0.3) and (6.95,1.4) .. (10.93,3.29)   ;
\draw    (194.8,231.5) -- (213.5,231.5) ;
\draw [shift={(215.5,231.5)}, rotate = 180] [color={rgb, 255:red, 0; green, 0; blue, 0 }  ][line width=0.75]    (10.93,-3.29) .. controls (6.95,-1.4) and (3.31,-0.3) .. (0,0) .. controls (3.31,0.3) and (6.95,1.4) .. (10.93,3.29)   ;
\draw    (440.8,160.5) -- (459.5,160.5) ;
\draw [shift={(461.5,160.5)}, rotate = 180] [color={rgb, 255:red, 0; green, 0; blue, 0 }  ][line width=0.75]    (10.93,-3.29) .. controls (6.95,-1.4) and (3.31,-0.3) .. (0,0) .. controls (3.31,0.3) and (6.95,1.4) .. (10.93,3.29)   ;
\draw  [fill={rgb, 255:red, 255; green, 255; blue, 255 }  ,fill opacity=1 ] (273.5,179.5) -- (338,179.5) -- (338,244) -- (273.5,244) -- cycle ;
\draw  [fill={rgb, 255:red, 192; green, 221; blue, 251 }  ,fill opacity=1 ] (273.5,179.5) -- (338,179.5) -- (338,202) -- (273.5,202) -- cycle ;
\draw  [fill={rgb, 255:red, 255; green, 255; blue, 255 }  ,fill opacity=1 ] (250.5,132.5) -- (315,132.5) -- (315,197) -- (250.5,197) -- cycle ;
\draw  [fill={rgb, 255:red, 192; green, 221; blue, 251 }  ,fill opacity=1 ] (251,132.5) -- (315,132.5) -- (315,155) -- (251,155) -- cycle ;
\draw    (333.8,157.5) -- (364,157.97) ;
\draw [shift={(366,158)}, rotate = 180.89] [color={rgb, 255:red, 0; green, 0; blue, 0 }  ][line width=0.75]    (10.93,-3.29) .. controls (6.95,-1.4) and (3.31,-0.3) .. (0,0) .. controls (3.31,0.3) and (6.95,1.4) .. (10.93,3.29)   ;
\draw  [fill={rgb, 255:red, 255; green, 255; blue, 255 }  ,fill opacity=1 ] (228.5,85.5) -- (293,85.5) -- (293,150) -- (228.5,150) -- cycle ;
\draw  [fill={rgb, 255:red, 192; green, 221; blue, 251 }  ,fill opacity=1 ] (228.5,85.5) -- (293,85.5) -- (293,108) -- (228.5,108) -- cycle ;

\draw (100,92) node [anchor=north west][inner sep=0.75pt]   [align=left] {\begin{minipage}[lt]{50.34pt}\setlength\topsep{0pt}
\begin{center}
\textbf{{AI Project}}
\end{center}

\end{minipage}};
\draw (465,136) node [anchor=north west][inner sep=0.75pt]   [align=left] {\begin{minipage}[lt]{52.71pt}\setlength\topsep{0pt}
\begin{center}
\textbf{{\scriptsize Compliance}}\\\textbf{{\scriptsize Determination }}\\\textbf{{\scriptsize for AI Project}}
\end{center}

\end{minipage}};
\draw (388,37) node [anchor=north west][inner sep=0.75pt]   [align=left] {\textbf{{\scriptsize {\fontfamily{pcr}\selectfont Step 2}}}};
\draw (234,37) node [anchor=north west][inner sep=0.75pt]   [align=left] {\textbf{{\scriptsize {\fontfamily{pcr}\selectfont Step 1}}}};
\draw (382,119) node [anchor=north west][inner sep=0.75pt]  [font=\normalsize] [align=left] {\begin{minipage}[lt]{33.04pt}\setlength\topsep{0pt}
\begin{center}
{\tiny \textbf{{Automated}}}\\{\tiny \textbf{{ compliance }}}\\{\tiny \textbf{{Analysis }}}\\{\tiny \textbf{{across all }}}\\{\tiny \textbf{{Compliance }}}\\{\tiny \textbf{{Cards}}}
\end{center}

\end{minipage}};
\draw (260,161.5) node [anchor=north west][inner sep=0.75pt]   [align=left] {\begin{minipage}[lt]{32.47pt}\setlength\topsep{0pt}
\begin{center}
{ \textbf{{\tiny Model CC(s)}}}
\end{center}

\end{minipage}};
\draw (286,207.5) node [anchor=north west][inner sep=0.75pt]   [align=left] {\begin{minipage}[lt]{28.79pt}\setlength\topsep{0pt}
\begin{center}
{\textbf{{\tiny Data CC(s)}}}
\end{center}

\end{minipage}};
\draw (237,112.5) node [anchor=north west][inner sep=0.75pt]   [align=left] {\begin{minipage}[lt]{28.79pt}\setlength\topsep{0pt}
\begin{center}
{\textbf{{\tiny Project CC}}}
\end{center}

\end{minipage}};
\draw (86,173) node [anchor=north west][inner sep=0.75pt]  [font=\normalsize] [align=left] {\begin{minipage}[lt]{28.79pt}\setlength\topsep{0pt}
\begin{center}
{\tiny \textbf{{Integrated }}}\\{\tiny \textbf{{ Models}}}
\end{center}

\end{minipage}};
\draw (143,173) node [anchor=north west][inner sep=0.75pt]  [font=\normalsize] [align=left] {\begin{minipage}[lt]{28.79pt}\setlength\topsep{0pt}
\begin{center}
{\tiny \textbf{{Integrated }}}\\{\tiny \textbf{{ Datasets}}}
\end{center}

\end{minipage}};

\end{tikzpicture}

\end{center}
\caption{\textit{\textbf{Compliance Cards AIA Compliance Analysis Procedure}: In Step 1, the Compliance Cards are used to collect, from internal and/or external sources, standardized metadata about the AI Project as well as its component Models and Datasets. In Step 2, an automated analysis looks across all of the gathered/harmonized metadata to render a compliance analysis for the overall AI Project.}}\label{fig:ccanalysis}
\end{figure}

\section{The Why and How of Compliance Cards}

The Compliance Cards system is shaped by three key features of the modern AI landscape. First, the Compiance Cards' exact metadata is informed by AI's emerging regulatory dimension--namely, the AIA's dual focus on both project- and component-level features of an AI project. Second, the Compliance Cards system's specialized, separate artifacts are a response to the growing complexity of the AI supply chain faced by AI providers. Lastly, the growing need for real-time compliance analyses in various provider workflows inspired the Compliance Cards Algorithm and the complimentary, computational format of the artifacts.

\subsection{The Dual Focus of the AIA's Requirements}

A key feature of the AIA (and perhaps other upcoming AI regulation\footnote{The White House's Blueprint for an AI Bill of Rights, for example, also champions rules that are aimed at both AI systems and models at-large \citep[19]{aibillofrights}, as well as their component datasets or models \citep[38]{aibillofrights}.}) is that some of its requirements target an AI project at-large, while others target the datasets or models that the project incorporates as components. As an example, the AIA's record-keeping requirements target a project at-large \citep[Art. 9, 12]{europa}. But the AIA's data governance requirement mostly target the datasets that a project incorporates as components \citep[Art. 10]{europa}.

Notably, these sets of requirements are heavily interconnected. For example, the exact requirements that the dataset and model components must satisfy are a function of several high-level characteristics of the AI project\footnote{We call this special tranche of high-level characteristics, which play a key role in our Compliance Cards Algorithm, ``dispositive characteristics.''}, including ---but not limited to-- whether the AI project consists of an AI system or a general-purpose AI (``GPAI'') model \citep[Art. 1(2), 2, 6]{europa}. Conversely, features of component models or datasets ---such as the interpretation tools built into models--- can shape what is required of the AI project at-large (e.g., the human oversight measures it must implement) \citep[Art. 14]{europa}.

The net effect of this duality ---and this interconnectedness--- is that providers cannot render a robust compliance analysis for an AI project without first gathering certain information about both the project at-large \emph{and} its component datasets and models. To facilitate this, we introduce novel transparency artifacts ---Compliance Cards, an evolution of earlier transparency artifacts such as Model Cards \citep{DBLP:conf/fat/MitchellWZBVHSR19}--- that are specifically designed to capture both kinds of essential information. 

\subsubsection{The growing complexity of the AI supply chain}

Where an AI project is housed entirely inside one team, gathering all of the information required to render a compliance analysis ---covering both the AI project at-large and any incorporated datasets or models--- may be trivial. But, as the AIA itself notes, ``[a]long the AI value chain multiple parties often supply ...components or processes that are incorporated'' into AI projects \citep[Rec. 88]{europa}. These components, often sourced externally \citep{DBLP:conf/icse/AmershiBBDGKNN019, DBLP:conf/emisa/TakeABSO21, googleresearch, Renieris2023-ah}, may include multiple\footnote{For example, Apple's OpenELM LLM was trained on four datasets \citep{DBLP:journals/corr/abs-2404-14619}, including The Pile, which is itself an aggregation of 22 datasets \citep{gao2020pile800gbdatasetdiverse}.} datasets and models obtained from open-source platforms like Hugging Face \citep{osborne2024ai, DBLP:journals/corr/abs-2406-08205} or third party models accessed via API \citep{adalovelaceinstituteWhatFoundation}. 
Notably, the provider integrating these components may lack insight into certain aspects of them (e.g., a model's training data) \citep{DBLP:conf/fat/LiesenfeldD24} and may never have direct contact with the supplier \citep{DBLP:journals/corr/abs-1907-03483}. 

In such scenarios, it may prove difficult to have all of the different component suppliers (along with the provider) collectively fill out a single artifact capturing all the information required for a compliance analysis. What may be more feasible is to have each of the parties fill out an artifact strictly focused on their own contribution, in a decentralized fashion, then for these separate artifacts to be united, Voltron-style, in order to render the compliance analysis. The advantages of this approach include: 

\begin{itemize}
    \item{Each party documents what they know best (i.e., their particular contribution to the aggregation), reaping the benefits of ``division of knowledge'' \citep[49]{hayek};} 
    \item{The artifacts can be filled-out (or updated) asynchronously, at each party's leisure;}
    \item{These artifacts can be efficiently shared in a one-to-many fashion (e.g., via Hugging Face);}
    \item{The parties do not need to be in contact in order for a compliance analysis to occur.\footnote{In this manner, Compliance Cards are ``MultiActor Responsible AI (RAI) Artifacts'' that can support direct or indirect communication between different stakeholders \citep{kawakami2024responsibleaiartifactsadvance}.}}
\end{itemize}

To reap these benefits, the Compliance Cards system was designed with three separate (yet compatible) types of artifacts ---Project, Data, and Model CCs--- that can be filled-out asynchronously by the parties best equipped to do so, then combined to provide a holistic view of an AI project when it is time to run a compliance analysis. 

\subsubsection{The need for real-time AI regulation compliance analyses in various AI workflows}\label{sec:workflows}

In a growing number of AI development scenarios, the ability to make real-time (or, at least, rapid) AI regulation compliance determinations would be highly advantageous for providers. These include:  

\paragraph{Today's rapidly iterative AI development workflows}Contemporary AI development is characterized by highly iterative experimentation that involves trying many different approaches ---including with different models and datasets--- to find the combination that performs best for the use case.\citep{carliniRapidIteration, 10.14778/3297753.3297763, DBLP:journals/corr/abs-2404-02081, DBLP:journals/corr/abs-2201-13224}. During this phase of ``continuous experimentation''\citep{DBLP:journals/tosem/Martinez-Fernandez22}, it would be beneficial for a provider to be able to check, in real-time, whether a candidate combination is regulation-compliant and, therefore, worth exploring.

\paragraph{Search and acquisition of components on AI communities and marketplaces} 
Finding the right model or dataset to integrate into an AI project may involve searching through many candidates on ML communities (e.g., Hugging Face), model zoos, and marketplaces \citep{DBLP:journals/vldb/ChapmanSKKIKG20, huggingfaceSuperchargedSearching}. Such searches often involve multiple queries that are refined in real-time based on incoming search results \citep{DBLP:conf/cikm/KimCTL13}. Real-time compliance analyses would allow providers, while searching for components, to instantly understand the compliance level of various candidate components as integrated into their overall AI project.

\paragraph{Federated learning} Federated learning (FL) is a privacy-preserving ML technique in which models learn from data without directly accessing it \citep{DBLP:conf/aistats/McMahanMRHA17}. FL involves updating a global model locally, on private data, then sharing those updates (rather than the data) back to a central server for aggregation into the global model. In this scenario, the  
central server, so as not to slow model convergence, can ideally make rapid decisions about whether or not integrating a given batch of client-side data (via model updates) is AIA-compliant. Real-time compliance analyses would support this.

\paragraph{Continuous integration / continuous delivery} 
There is great interest in automating the retraining of in-production AI models using production data, a process known as continuous integration and continuous delivery \citep{DBLP:journals/corr/abs-2202-03541, DBLP:conf/kdd/KarlasIR0ZBELXW20}. In such scenarios, it would be advantageous for a provider to know, in real-time, whether the integration of an incoming batch of new data (and/or the model retrained on that data), when combined with the overall AI project, are regulation-compliant. 

The need for rapid or real-time compliance analyses in these AI development situations and others led to the design requirement that Compliance Cards must be readily manipulable by some accompanying algorithm in order to render low-latency compliance analyses. This, in turn, inspired the computational format of Compliance Cards ---as well as the complimentary Compliance Cards Algorithm.
 
\section{The Compliance Cards System}

The Compliance Cards system has two key elements: (1) the Compliance Cards transparency artifacts; and (2) the Compliance Cards Algorithm. In this section, we take a closer look at each. 

\subsection{Compliance Cards}

The Compliance Cards are an interlocking set of computational transparency artifacts that capture compliance-related metadata about both an AI project as well as its constituent datasets and models. Let's examine their overall format, then describe each type of card in more detail:   

\subsubsection{Compliance Cards format}

A defining quality of Compliance Cards is that their metadata can be operated on by a Compliance Cards Algorithm to produce a compliance analysis. While there are different ways to embody this quality, they all follow the same overarching structure: a set of attribute-value pairs. These can inhabit many formats, including via human-readable data serialization language (e.g., YAML or JSON) or serialized python objects like dictionaries. In each of these, the attributes will capture specific characteristics of the subject (whether it is a project, model, or dataset) that relate to AIA compliance, while the values will quantify whether or how much the subject possesses that attribute. In the form we deem optimal, the latter are quantized ---i.e., constrained to a finite set consisting of integers (e.g., confidence scores), boolean values, or other. Because it is the simplest format for computational manipulation, this is what we use in the initial, YAML-based Compliance Card templates\footnote{Our templates are composed in YAML so that they can readily be added to the existing YAML metadata sections of the (popular) Hugging Face Model Cards, lowering the friction of adoption \citep{githubHubdocsmodelcardmdyaml}.} that are available at: \texttt{https://github.com/camlsys/compliancecards}.


\subsubsection{Compliance Cards types}

There are three types of Compliance Card (``CC'')  artifacts:

\paragraph{Project CC} This type of Compliance Card captures AIA-relevant attributes of an AI project, including the ``dispositive attributes'' that heavily inform the compliance analysis. Table~\ref{table:project} presents an overview of the attributes this artifact captures. For a full list, see the \texttt{project\_cc.yaml} template in the GitHub repository for this project: \texttt{https://github.com/camlsys/compliancecards}.


\paragraph{Data CC} This type of Compliance Card captures AIA-relevant attributes of a component dataset (whether training, evaluation, or test set). Table~\ref{table:data} presents an overview of the attributes it captures. For a full list, see the \texttt{data\_cc.yaml} template in the GitHub repository for this project.

\paragraph{Model CC} This type of Compliance Card captures AIA-relevant attributes of a component model (including general-purpose AI models\footnote{The AIA highlights that GPAI models ``can be integrated as a component of a downstream AI system covered by the AIA, but also...can be used `as such' as an AI system falling under the scope of the [AIA]'' \citep{Fernández-Llorca2024}.}). Table~\ref{table:model} presents an overview of the attributes it captures. For a full list, see the \texttt{model\_cc.yaml} template in the GitHub repository for this project.

Note that an AI project integrating multiple models or datasets will have a one-to-many relationship between its Project CC and/or Data and Model CCs.

\subsection{Compliance Cards Algorithm}

The Compliance Cards Algorithm looks across all of the metadata in the Compliance Cards to render a real-time prediction about whether the aggregate AI project complies with the AIA (Figure~\ref{fig:ccalgorithm}).  

\begin{figure}[ht]\label{fig:ccalgorithm}

\begin{center}

\scalebox{0.7}{

\tikzset{every picture/.style={line width=0.75pt}} 

\begin{tikzpicture}[x=0.75pt,y=0.75pt,yscale=-1,xscale=1]

\draw  [fill={rgb, 255:red, 255; green, 255; blue, 255 }  ,fill opacity=1 ] (168.5,197.5) -- (233,197.5) -- (233,262) -- (168.5,262) -- cycle ;
\draw  [fill={rgb, 255:red, 192; green, 221; blue, 251 }  ,fill opacity=1 ] (168.5,197.5) -- (233,197.5) -- (233,220) -- (168.5,220) -- cycle ;

\draw  [fill={rgb, 255:red, 255; green, 255; blue, 255 }  ,fill opacity=1 ] (145.5,150.5) -- (210,150.5) -- (210,215) -- (145.5,215) -- cycle ;
\draw  [fill={rgb, 255:red, 192; green, 221; blue, 251 }  ,fill opacity=1 ] (146,150.5) -- (210,150.5) -- (210,173) -- (146,173) -- cycle ;
\draw    (219.8,176.5) -- (250,176.97) ;
\draw [shift={(252,177)}, rotate = 180.89] [color={rgb, 255:red, 0; green, 0; blue, 0 }  ][line width=0.75]    (10.93,-3.29) .. controls (6.95,-1.4) and (3.31,-0.3) .. (0,0) .. controls (3.31,0.3) and (6.95,1.4) .. (10.93,3.29)   ;
\draw  [fill={rgb, 255:red, 255; green, 255; blue, 255 }  ,fill opacity=1 ] (123.5,103.5) -- (188,103.5) -- (188,168) -- (123.5,168) -- cycle ;
\draw  [fill={rgb, 255:red, 192; green, 221; blue, 251 }  ,fill opacity=1 ] (123.5,103.5) -- (188,103.5) -- (188,126) -- (123.5,126) -- cycle ;

\draw    (446.8,173.5) -- (477,173.97) ;
\draw [shift={(479,174)}, rotate = 180.89] [color={rgb, 255:red, 0; green, 0; blue, 0 }  ][line width=0.75]    (10.93,-3.29) .. controls (6.95,-1.4) and (3.31,-0.3) .. (0,0) .. controls (3.31,0.3) and (6.95,1.4) .. (10.93,3.29)   ;
\draw  [fill={rgb, 255:red, 189; green, 16; blue, 224 }  ,fill opacity=0.32 ] (261,134) -- (278,117) -- (443,117) -- (443,219) -- (426,236) -- (261,236) -- cycle ; \draw   (443,117) -- (426,134) -- (261,134) ; \draw   (426,134) -- (426,236) ;

\draw (486,150) node [anchor=north west][inner sep=0.75pt]   [align=left] {\begin{minipage}[lt]{74.13pt}\setlength\topsep{0pt}
\begin{center}
\textbf{Compliance}\\\textbf{Determination }\\\textbf{for AI Project}
\end{center}

\end{minipage}};
\draw (157,179.5) node [anchor=north west][inner sep=0.75pt]   [align=left] {\begin{minipage}[lt]{26.24pt}\setlength\topsep{0pt}
\begin{center}
{\textbf{{\tiny Model CC}}}
\end{center}

\end{minipage}};
\draw (282,164) node [anchor=north west][inner sep=0.75pt]  [font=\normalsize] [align=left] {\begin{minipage}[lt]{86.63pt}\setlength\topsep{0pt}
\begin{center}
Compliance Cards\\Algorithm
\end{center}

\end{minipage}};
\draw (132,130.5) node [anchor=north west][inner sep=0.75pt]   [align=left] {\begin{minipage}[lt]{28.79pt}\setlength\topsep{0pt}
\begin{center}
{\textbf{{\tiny Project CC}}}
\end{center}

\end{minipage}};
\draw (184,225.5) node [anchor=north west][inner sep=0.75pt]   [align=left] {\begin{minipage}[lt]{22.56pt}\setlength\topsep{0pt}
\begin{center}
{\textbf{{\tiny Data CC}}}
\end{center}

\end{minipage}};

\end{tikzpicture}

}

\end{center}
\caption{\textit{\textbf{The Compliance Cards Algorithm}: The Compliance Cards Algorithm is any algorithm that accepts the Compliance Cards as its input and outputs a compliance analysis.}}
\end{figure}

While the Compliance Cards Algorithm can take on many different embodiments, the initial embodiment that we put forward in conjunction with this paper is rules-based and uses a series of if-then statements to represent and encode the requirements of the the AIA. A Python implementation of this algorithm is available at the Compliance Cards GitHub repository. Importantly, other embodiments are not only possible but may scale better to unseen AI projects. In Section~\ref{sec:future}, for example, we express our intent to produce an LLM-assisted embodiment of the Compliance Cards Algorithm. 

\section{How to Use the Compliance Cards System}

How should the Compliance Cards system actually be used in practice? Here are some ideas: 

\subsection{Who should populate the Compliance Cards?}

In light of the information asymmetry that exists along the AI supply chain \citep{DBLP:journals/corr/abs-1907-03483}, Compliance Cards are ideally populated by whichever entity has the best information about the subject of the respective card. As a rule of thumb, we recommend the entities in Table~\ref{table:cc-entities} populate each artifact.\footnote{Note that the AIA explicitly encourages AI component makers to provide documentation to the integrators of those components ---something we argue is achieved by sharing filled-out Compliance Cards  \cite[Rec. 88, 89]{europa}.}

\begin{table}[h]
  \caption{Who should populate the Compliance Card?}
  \label{table:cc-entities}
  \centering
  \begin{tabular}{p{3cm} p{5cm}}
    \toprule
    \cmidrule(r){1-2}
    Artifact     & Recommended populator \\
    \midrule
Project CC &  Provider    \\
\hline
Data CC &  Dataset Supplier(s)
    \\
\hline
Model CC & Model Supplier(s)    \\
    \bottomrule
  \end{tabular}
\end{table}

\subsection{How should the Compliance Cards be populated?}

While parties can populate their Compliance Cards any way they like, here are some methods that we expect to be popular: 

\begin{itemize}
    \item{\textbf{Manual population via user interface (UI)}: Compliance Cards can be populated manually through a questionnaire-style UI akin to those used to populate existing artifacts  \citep{huggingfaceModelcardCreator}. In the case of Compliance Cards, a UI like this might present each individual attribute to the user, asking them to enter the appropriate value for that attribute.}
    \item{\textbf{Programmatic population via library}: Like some existing transparency artifacts, Compliance Cards could hypothetically be populated programmatically, via a library that is referenced from within model training, evaluation, and deployment code  \citep{huggingfaceModelCards, pypiModelcardtoolkit, researchIntroducingModel}. In Section~\ref{sec:future}, we express our intent to release a similar library for Compliance Cards.} 
    \item{\textbf{Population via large language model (LLM)}: Text documents, including but not limited to training and evaluation code, data governance policies, or other artifacts such as Model and Data Cards, could be inputted into a specialized LLM that outputs one or more populated Compliance Cards. In Section~\ref{sec:future}, we express our intent to release a toolkit that allows users to leverage existing LLMs to populate Compliance Cards in this manner.} 
\end{itemize}

\subsection{What happens to Compliance Cards after they are populated?}

Our hope is that Data and Model CCs, like existing transparency artifacts, will be uploaded to dataset and model repositories on ML communities and marketplaces (e.g., Hugging Face). There, they can be readily accessed by providers or other parties (discussed in Section~\ref{sec:who}) who may wish to combine them with a Project CC and/or other Data and Model CCs, then render a compliance analysis for a potential or existing AI project. Alternatively, the suppliers of datasets and models may wish to generate these artifacts and keep them on-hand for private transmission to  such parties. 

Because ML communities and marketplaces are not yet, as far as we know, instrumented to host Project CCs in the same way they can host Data and Model CCs, we expect that the Project CC for a given AI project is more likely to be generated by an AI provider and then kept locally until that provider (or any of the parties listed in Section~\ref{sec:who}) wishes to pair it with any relevant Data and Model CCs and render a compliance analysis for the project.

\subsubsection{Who will run the Compliance Cards Algorithm on the Compliance Cards?}\label{sec:who}

In short, anyone with the full set of Compliance Cards associated with an AI project can run the Compliance Cards Algorithm on them and output a compliance analysis. That said, here are some users and use cases we expect to be commonplace: 

\begin{itemize}
\item{\textbf{Conformity assessors}: The AIA requires AI providers and other operators, before putting certain AI projects on the EU market, to undergo a conformity assessment ---performed by themselves or a third party--- verifying the project's AIA-compliance \citep[Art. 16, 43]{europa}. We believe the Compliance Cards system can accelerate these assessments, regardless of who performs them.} 
\item{\textbf{Market surveillance entities}: Under the AIA, market surveillance authorities have the power to investigate AIA non-compliance \citep[Art. 75]{europa}. We believe the Compliance Cards system can accelerate such investigations as regards providers.\footnote{AI regulators may suffer from a comparative lack of capacity, resources, and expertise \citep{DBLP:journals/corr/abs-2404-09932}. By helping overcome this, the Compliance Cards may help boost AIA enforcement.}}  
\item{\textbf{Providers}: Because providers of AI projects not only bear the most responsibility for satisfying the AIA's requirements \citep[Arts. 2(1), 3(3)]{europa, edri}, but also own the development workflows (Section~\ref{sec:workflows}) that would most benefit from accelerated compliance analyses, we anticipate they could make the Compliance Cards system a key part of the internal assessments that are ``the core of compliance with the [AIA]'' \citep{DBLP:conf/fat/LucajSB23, DBLP:journals/aiethics/ThelissonV24}.}

\item{\textbf{Dataset and model suppliers}: We envision a scenario in which model and dataset suppliers use the Compliance Cards system to evaluate the compliance level of their wares when paired with various providers' Project CCs, using the output to inform everything from their future development plans to their marketing strategy.}
\end{itemize}

\subsubsection{How will they run the Compliance Cards Algorithm on the Compliance Cards?}

Our Python implementation of the Compliance Cards Algorithm is open source and can be used by anyone interested in rendering a compliance analysis for an AI project. Our hope is that the algorithm will be integrated into various applications or platforms, made available to their users to democratize its use. In the meantime, we also plan to host the algorithm in a web application that anyone can use to run analyses across a set of Compliance Cards they generate and/or upload. 

\section{Related Work}

Researchers have developed various transparency artifacts to facilitate the sharing of metadata about models, datasets, and other facets of AI systems. Some of these artifacts have been popularized by industry. Most recently, there have been efforts to produce transparency artifacts that are machine-readable as well as efforts to produce transparency artifacts specialized to AI regulation, including the AIA. In parallel, there has been work on algorithms that encode the requirements of the AIA. What is still missing ---and what Compliance Cards provides--- is a holistic system that brings together all the necessary building blocks for real-time compliance analyses by pairing computational, AIA-specialized transparency artifacts with a connected, AIA-encoding algorithm.

\subsection{Human- and machine-readable transparency artifacts}

Multiple efforts have produced human-readable transparency artifacts that facilitate the sharing of metadata about AI components. This includes artifacts for
datasets \citep{DBLP:journals/tacl/BenderF18, 10.1145/3458723, DBLP:conf/fat/MitchellWZBVHSR19, Holland2018TheDN, DBLP:journals/corr/abs-2201-03954, Hutchinson2020TowardsAF, 10.1145/3531146.3533239}, models \citep{DBLP:conf/fat/MitchellWZBVHSR19, researchIntroducingModel, tensorflowModelCard,DBLP:conf/cogmi/SeifertSW19, 10.1145/3442188.3445971, DBLP:journals/corr/abs-2204-13582, ibmUsingFactsheets}, services \citep{DBLP:journals/ibmrd/ArnoldBHHMMNROP19}, and multi-model systems  \citep{metaSystemLevelTransparency}. 

Recognizing the need for transparency artifacts that were more readily analyzable by computers, researchers have recently captured various transparency artifacts in machine-readable formats like the OWL 2 Web Ontology Language \citep{DBLP:journals/bmcbi/AmithCZRJLYT22}, linked data and knowledge graphs \citep{10.1145/3543873.3587659}, YAML
\citep{booth}, and JSON \citep{DBLP:journals/corr/abs-2312-06153}. Perhaps most notable due to their widespread use\footnote{Over 32,000 Model Cards and 7000 Dataset Cards have been filled-out and shared on Hugging Face \citep{liang2024whats, DBLP:journals/corr/abs-2401-13822}.}, Hugging Face Model Cards are captured as machine-readable Markdown files with optional YAML sections \citep{githubHuggingface_markdown, githubHubdocsmodelcardmdyaml}, a format that enabled a large-scale analysis of these Model Cards' contents by \citet{liang2024whats}.

\subsection{Transparency artifacts for AI regulation (including AIA) compliance}

Although the transparency artifacts above capture some of the metadata needed for an AIA compliance determination \citep{DBLP:conf/aisafety/CalanzoneCTOC23}, they do not capture all of it. To address this gap, some recent efforts have introduced transparency artifacts specialized to the AIA compliance (or, at least, AI regulation compliance) use case. In particular, \citet{DBLP:journals/corr/abs-2307-11525} introduced human-readable transparency artifacts to help align AI systems with AI regulations during development while \citet{DBLP:journals/ethicsit/HupontLBG24} introduced human-readable Use Case Cards to capture metadata related to Act's risk levels. \citet{DBLP:conf/caise/LohachabU24} rendered model artifacts as blockchain NFTs to, among other things, advance regulatory compliance. Most recently,
\citet{golpayegani2024aicardsappliedframework} produced a transparency artifact that captures metadata related to the corner of the AIA addressing AI risk management technical documentation. Importantly, none of these efforts include a comprehensive (and certainly not automated) method for looking across their metadata to produce a compliance analysis.

\subsection{AIA-encoding Algorithms}

RegCheckAI was a web application, apparently discontinued, that claimed to render an AIA compliance determination based solely on the metadata in a Hugging Face Model Card (an approach we argue is incomplete for many AI projects) \citep{huggingfaceregcheck}. Meanwhile, \citet{ComplianceChecker} offers a web application that predicts the AIA requirements that may apply to AI project based on its inputted details; importantly, it does not analyze whether the project satisfies those requirements. A handful of startups also offer web applications advertising similar functionality, sometimes behind registration walls \citep{trailmlComplianceChecker, holisticaiRiskCalculator}. While the algorithms undergirding these web applications are not public, we infer that they include some type of encoding of the AIA's rules (or, at least, a portion of them).

\subsection{The void that remains (and Compliance Cards fills)}

What has been sorely missing from existing research is a holistic system that interweaves all of the the necessary building blocks for real-time AIA compliance analyses: (1) Computation-amenable artifacts that captures all of the metadata necessary for a robust AIA compliance analysis; (2) A companion algorithm that operates on those artifacts to output a 360 AIA compliance analysis. The Compliance Cards system is meant to fill this void. 

\section{Future Work}\label{sec:future}

There are many opportunities to extend this line of research, with the Compliance Cards system as a foundation:

\subsection{Evaluation}

It would be useful to thoroughly evaluate the accuracy, speed, and expense of the Compliance Cards system as compared to the status quo (the manual compliance analyses depicted in Figure~\ref{fig:statusquo}). 

\subsection{Automate the process of Compliance Card population}
Automating the population of Compliance Cards will further streamline the process of AIA compliance analyses. There are multiple paths towards this goal: 

\begin{itemize}
\item{Values could be programmatically deposited into Compliance Cards during the training and evaluation processes. This could include the outputs of existing responsible AI benchmarking frameworks \citep{DBLP:conf/acl/LinHE22, DBLP:conf/acl/ParrishCNPPTHB22}. This process could be managed by a library that bears similarity to the existing artifact-populating libraries like the Model Cards Toolkit of~\citep{researchIntroducingModel, tensorflowModelCard, amazonModelCard}.}
\item{Training, evaluation, and product code, product specifications, existing transparency artifacts, and other documentation can be inputted into an LLM that, perhaps via contextual learning, outputs populated Compliance Cards.} 
\end{itemize}

\subsection{Provide tools to verify Compliance Cards metadata}

Sometimes, component models or datasets will not be accessible to their  integrators or would-be integrators; this may happen wherever these components are closed by design \citep{DBLP:conf/eacl/BalloccuSLD24}, before their exchange during arms-length business transactions \citep{DBLP:journals/tdsc/WengWCHW22}, or in federated learning \citep{woisetschläger2024federatedlearningprioritieseuropean}. In such scenarios, it would nonetheless be valuable for model and dataset integrators to be able to verify that the metadata shared in Model and Data Compliance Cards is accurate \citep{lcfiActsTechnical, reuel2024openproblemstechnicalai}. There has been promising work using zk-SNARK zero-knowledge proofs to verify ML model accuracy and fairness without direct model access \citep{DBLP:journals/corr/abs-2402-02675}, though it may come with high latency. Differently, Hugging Face has unveiled a feature that lets model makers run evaluations on Hugging Face-controlled servers and receive metrics, digitally signed by Hugging Face, to add to their Model Cards \citep{huggingfaceFacebookbartlargecnnVerifyToken}. It would be valuable to explore how these or other closed-box methods can be used to verify the metadata captured in Compliance Cards (in a cost-effective and rapid manner). 

\subsection{Adapt Compliance Cards to AIA technical standards and other AI regulation}
As is typical in EU lawmaking, multiple EU standardization bodies have been asked to create harmonized standards ("common specifications") to accompany the AIA \citep{cencenelec, europaDocs, Wachter}. AI projects that conform to these "more concrete" \citep{DBLP:journals/corr/abs-2208-12645} requirements will be presumed to be AIA-compliant \citep[41(3)]{europa}. Once issued, there could be value in replacing some or all of the existing Compliance Card attributes with ones that reflect these more granular specifications. 

In addition, other AI regulations are being gestated worldwide \citep{reuterslatest}. The Compliance Cards framework could be extended beyond the AIA and to these other regulations as they materialize.

\section{Conclusion}

The Compliance Card system is designed to reduce the burden of AIA compliance analyses that firms face admidst an increasingly complex AI supply chain. Our hope (and hunch) is that lowering the capacity required to run AIA compliance analyses will give AI firms (especially resource-starved small- and medium-sized businesses) more bandwidth to cure any of the deficiencies these analyses uncover or to investigate nuanced issues that an automated analysis may not fully capture. Accordingly, we see Compliance Cards as a pathway to increased AIA compliance and, since the AIA encodes so many core responsible AI principles, a step forward for responsible AI in general. 

\bibliographystyle{plainnat}
\bibliography{neurips_2024}


\appendix
\section{Tables}

\begin{table}[H]
  \caption{Project CC Attributes (Overview)} \label{table:project}
  \centering
  \begin{tabular}{|p{7cm}|p{5cm}|}
    \toprule
    \cmidrule(r){1-2}
    Attribute category (dispositive characteristics in italics)     & Relevant Section(s) of AIA \\
    \midrule
\textit{Operator role (provider)} &  Art. 2   \\
\hline
\textit{Intended purpose} &  Art. 6, 9, 10, 13, 14, 15   \\
\hline
\textit{Whether on EU market}  &  Art. 2   \\
\hline
\textit{Whether AI System, High-risk AI System, GPAI model, GPAI model with Systemic Risk} & Art. 3(1), 3(63), 6   \\
\hline
Whether excepted from AIA &  Art. 2(6)    \\
\hline
Whether prohibited practices  & Art. 5   \\
\hline
Risk management (High-risk AI systems)  &  Art. 9    \\
\hline
Data and data governance (High-risk AI systems)  & Art. 10   \\
\hline
Technical documentation (High-risk AI systems and GPAI models))  & Art. 11, 53(1); Annex XI(2), XI(1)   \\
\hline
Record-keeping (High-risk AI systems)  & Art. 12   \\
\hline
Transparency and provision of information to deployers (High-risk AI systems)   & Art. 13   \\
\hline
Human oversight (High-risk AI systems)  & Art. 14    \\
\hline
Accuracy, robustness, and cybersecurity (High-risk AI systems)  & Art.  15    \\
\hline
Registration, etc. (AI systems)  & Art.  16    \\
\hline
Fundamental Rights Assessment (High-risk AI systems)  & Art.  27    \\
\hline
Transparency (AI systems)  & Art.  50    \\
    \bottomrule
  \end{tabular}
\end{table}

\begin{table}[H]
  \caption{Data CC Attributes (Overview)}
  \label{table:data}
  \centering
  \begin{tabular}{|p{7cm}|p{5cm}|}
    \toprule
    \cmidrule(r){1-2}
    Attribute category (dispositive characteristics in italics)     & Relevant Section(s) of AIA \\
    \midrule
    \textit{Intended purpose} &  Art. 6, 9, 10, 13, 14, 15    \\
\hline
    Data and data governance (High-risk AI systems) & Art. 10; Rec. 10, 67     \\
\hline
Technical documentation (High-risk AI systems and GPAI models) & Art. 11, 13, 53(1); Annex IV, XI; \\
\hline
Accuracy, robustness and cybersecurity (High-risk AI systems)  & Art. 15     \\
\hline
Quality management system (High-risk AI systems) & Art. 17(1)(f) \\
    \bottomrule
  \end{tabular}
\end{table}

\begin{table}[H]
  \caption{Model CC Attributes (Overview)}
  \label{table:model}
  \centering
  \begin{tabular}{|p{7cm}|p{5cm}|}
    \toprule
    \cmidrule(r){1-2}
    Attribute category (dispositive characteristics in italics)      & Relevant Section(s) of AIA \\
    \midrule
\textit{Intended purpose} &  Art. 6, 9, 10, 13, 14, 15    \\
\hline
Risk management (High-risk AI systems) &  Art. 9(2), 9(6)    \\
\hline
Data and data governance (High-risk AI systems) &  Art. 10    \\
\hline
Technical documentation (High-risk AI systems and GPAI models) & Art. 11, 53(1); Annex IV, XI; \\
\hline
Transparency and provision of information to deployers (High-risk AI systems, GPAI models, and GPAI models with systemic risk) & Art. 13    \\
\hline
Human oversight (High-risk AI systems) & Art. 14(3)(a), 14(4)(d)    \\
\hline
Accuracy, robustness and cybersecurity (High-risk AI systems)   & Art. 15     \\
    \bottomrule
  \end{tabular}
\end{table}

\section{Diagrams}
\subsection{Compliance Cards Algorithm Flowchart}

\begin{center}

\scalebox{0.7}{

\tikzset{every picture/.style={line width=0.75pt}} 

\begin{tikzpicture}[x=0.75pt,y=0.75pt,yscale=-1,xscale=1]

\draw  [fill={rgb, 255:red, 182; green, 236; blue, 252 }  ,fill opacity=1 ] (457.5,26) -- (559.5,26) -- (559.5,74.84) .. controls (495.75,74.84) and (508.5,92.45) .. (457.5,81.06) -- cycle ; \draw  [fill={rgb, 255:red, 182; green, 236; blue, 252 }  ,fill opacity=1 ] (444.75,33.4) -- (546.75,33.4) -- (546.75,82.24) .. controls (483,82.24) and (495.75,99.85) .. (444.75,88.46) -- cycle ; \draw  [fill={rgb, 255:red, 182; green, 236; blue, 252 }  ,fill opacity=1 ] (432,40.8) -- (534,40.8) -- (534,89.64) .. controls (470.25,89.64) and (483,107.25) .. (432,95.86) -- cycle ;
\draw    (495.5,93) -- (495.5,128) ;
\draw [shift={(495.5,130)}, rotate = 270] [color={rgb, 255:red, 0; green, 0; blue, 0 }  ][line width=0.75]    (10.93,-3.29) .. controls (6.95,-1.4) and (3.31,-0.3) .. (0,0) .. controls (3.31,0.3) and (6.95,1.4) .. (10.93,3.29)   ;
\draw  [fill={rgb, 255:red, 236; green, 185; blue, 247 }  ,fill opacity=1 ] (434,133) -- (558.5,133) -- (558.5,208) -- (434,208) -- cycle ;
\draw  [fill={rgb, 255:red, 242; green, 237; blue, 171 }  ,fill opacity=1 ] (664.5,246) -- (724.5,283) -- (664.5,320) -- (604.5,283) -- cycle ;
\draw    (495.5,209) -- (495.5,244) ;
\draw [shift={(495.5,246)}, rotate = 270] [color={rgb, 255:red, 0; green, 0; blue, 0 }  ][line width=0.75]    (10.93,-3.29) .. controls (6.95,-1.4) and (3.31,-0.3) .. (0,0) .. controls (3.31,0.3) and (6.95,1.4) .. (10.93,3.29)   ;
\draw    (722.5,283) -- (771.5,283) ;
\draw [shift={(773.5,283)}, rotate = 180] [color={rgb, 255:red, 0; green, 0; blue, 0 }  ][line width=0.75]    (10.93,-3.29) .. controls (6.95,-1.4) and (3.31,-0.3) .. (0,0) .. controls (3.31,0.3) and (6.95,1.4) .. (10.93,3.29)   ;
\draw    (664.5,320) -- (664.5,355) ;
\draw [shift={(664.5,357)}, rotate = 270] [color={rgb, 255:red, 0; green, 0; blue, 0 }  ][line width=0.75]    (10.93,-3.29) .. controls (6.95,-1.4) and (3.31,-0.3) .. (0,0) .. controls (3.31,0.3) and (6.95,1.4) .. (10.93,3.29)   ;
\draw  [fill={rgb, 255:red, 196; green, 246; blue, 142 }  ,fill opacity=1 ] (784.7,263) -- (832.3,263) .. controls (838.49,263) and (843.5,271.95) .. (843.5,283) .. controls (843.5,294.05) and (838.49,303) .. (832.3,303) -- (784.7,303) .. controls (778.51,303) and (773.5,294.05) .. (773.5,283) .. controls (773.5,271.95) and (778.51,263) .. (784.7,263) -- cycle ;
\draw    (725.5,394) -- (774.5,394) ;
\draw [shift={(776.5,394)}, rotate = 180] [color={rgb, 255:red, 0; green, 0; blue, 0 }  ][line width=0.75]    (10.93,-3.29) .. controls (6.95,-1.4) and (3.31,-0.3) .. (0,0) .. controls (3.31,0.3) and (6.95,1.4) .. (10.93,3.29)   ;
\draw  [fill={rgb, 255:red, 196; green, 246; blue, 142 }  ,fill opacity=1 ] (787.7,374) -- (835.3,374) .. controls (841.49,374) and (846.5,382.95) .. (846.5,394) .. controls (846.5,405.05) and (841.49,414) .. (835.3,414) -- (787.7,414) .. controls (781.51,414) and (776.5,405.05) .. (776.5,394) .. controls (776.5,382.95) and (781.51,374) .. (787.7,374) -- cycle ;
\draw    (664.5,432) -- (664.5,467) ;
\draw [shift={(664.5,469)}, rotate = 270] [color={rgb, 255:red, 0; green, 0; blue, 0 }  ][line width=0.75]    (10.93,-3.29) .. controls (6.95,-1.4) and (3.31,-0.3) .. (0,0) .. controls (3.31,0.3) and (6.95,1.4) .. (10.93,3.29)   ;
\draw    (723.5,506) -- (772.5,506) ;
\draw [shift={(774.5,506)}, rotate = 180] [color={rgb, 255:red, 0; green, 0; blue, 0 }  ][line width=0.75]    (10.93,-3.29) .. controls (6.95,-1.4) and (3.31,-0.3) .. (0,0) .. controls (3.31,0.3) and (6.95,1.4) .. (10.93,3.29)   ;
\draw  [fill={rgb, 255:red, 196; green, 246; blue, 142 }  ,fill opacity=1 ] (785.7,486) -- (833.3,486) .. controls (839.49,486) and (844.5,494.95) .. (844.5,506) .. controls (844.5,517.05) and (839.49,526) .. (833.3,526) -- (785.7,526) .. controls (779.51,526) and (774.5,517.05) .. (774.5,506) .. controls (774.5,494.95) and (779.51,486) .. (785.7,486) -- cycle ;
\draw    (664.5,540) -- (664.5,575) ;
\draw [shift={(664.5,577)}, rotate = 270] [color={rgb, 255:red, 0; green, 0; blue, 0 }  ][line width=0.75]    (10.93,-3.29) .. controls (6.95,-1.4) and (3.31,-0.3) .. (0,0) .. controls (3.31,0.3) and (6.95,1.4) .. (10.93,3.29)   ;
\draw    (721.5,614) -- (770.5,614) ;
\draw [shift={(772.5,614)}, rotate = 180] [color={rgb, 255:red, 0; green, 0; blue, 0 }  ][line width=0.75]    (10.93,-3.29) .. controls (6.95,-1.4) and (3.31,-0.3) .. (0,0) .. controls (3.31,0.3) and (6.95,1.4) .. (10.93,3.29)   ;
\draw    (606,614) -- (543.5,614) ;
\draw [shift={(541.5,614)}, rotate = 360] [color={rgb, 255:red, 0; green, 0; blue, 0 }  ][line width=0.75]    (10.93,-3.29) .. controls (6.95,-1.4) and (3.31,-0.3) .. (0,0) .. controls (3.31,0.3) and (6.95,1.4) .. (10.93,3.29)   ;
\draw    (893.5,614) -- (942.5,614) ;
\draw [shift={(944.5,614)}, rotate = 180] [color={rgb, 255:red, 0; green, 0; blue, 0 }  ][line width=0.75]    (10.93,-3.29) .. controls (6.95,-1.4) and (3.31,-0.3) .. (0,0) .. controls (3.31,0.3) and (6.95,1.4) .. (10.93,3.29)   ;
\draw    (832.5,651) -- (832.5,686) ;
\draw [shift={(832.5,688)}, rotate = 270] [color={rgb, 255:red, 0; green, 0; blue, 0 }  ][line width=0.75]    (10.93,-3.29) .. controls (6.95,-1.4) and (3.31,-0.3) .. (0,0) .. controls (3.31,0.3) and (6.95,1.4) .. (10.93,3.29)   ;
\draw  [fill={rgb, 255:red, 242; green, 237; blue, 171 }  ,fill opacity=1 ] (664.5,357) -- (724.5,394) -- (664.5,431) -- (604.5,394) -- cycle ;
\draw  [fill={rgb, 255:red, 242; green, 237; blue, 171 }  ,fill opacity=1 ] (664.5,469) -- (724.5,506) -- (664.5,543) -- (604.5,506) -- cycle ;
\draw  [fill={rgb, 255:red, 236; green, 185; blue, 247 }  ,fill opacity=1 ] (945,574) -- (1069.5,574) -- (1069.5,649) -- (945,649) -- cycle ;
\draw  [fill={rgb, 255:red, 242; green, 237; blue, 171 }  ,fill opacity=1 ] (664.5,577) -- (724.5,614) -- (664.5,651) -- (604.5,614) -- cycle ;
\draw  [fill={rgb, 255:red, 242; green, 237; blue, 171 }  ,fill opacity=1 ] (481.5,577) -- (541.5,614) -- (481.5,651) -- (421.5,614) -- cycle ;
\draw  [fill={rgb, 255:red, 242; green, 237; blue, 171 }  ,fill opacity=1 ] (832.5,577) -- (892.5,614) -- (832.5,651) -- (772.5,614) -- cycle ;
\draw    (1010.5,649) -- (1010.5,684) ;
\draw [shift={(1010.5,686)}, rotate = 270] [color={rgb, 255:red, 0; green, 0; blue, 0 }  ][line width=0.75]    (10.93,-3.29) .. controls (6.95,-1.4) and (3.31,-0.3) .. (0,0) .. controls (3.31,0.3) and (6.95,1.4) .. (10.93,3.29)   ;
\draw    (421.5,614) -- (372.5,614) ;
\draw [shift={(370.5,614)}, rotate = 360] [color={rgb, 255:red, 0; green, 0; blue, 0 }  ][line width=0.75]    (10.93,-3.29) .. controls (6.95,-1.4) and (3.31,-0.3) .. (0,0) .. controls (3.31,0.3) and (6.95,1.4) .. (10.93,3.29)   ;
\draw    (481.5,651) -- (481.5,686) ;
\draw [shift={(481.5,688)}, rotate = 270] [color={rgb, 255:red, 0; green, 0; blue, 0 }  ][line width=0.75]    (10.93,-3.29) .. controls (6.95,-1.4) and (3.31,-0.3) .. (0,0) .. controls (3.31,0.3) and (6.95,1.4) .. (10.93,3.29)   ;
\draw  [fill={rgb, 255:red, 236; green, 185; blue, 247 }  ,fill opacity=1 ] (243,576) -- (367.5,576) -- (367.5,651) -- (243,651) -- cycle ;
\draw    (951.5,724) -- (901,724) ;
\draw [shift={(899,724)}, rotate = 360] [color={rgb, 255:red, 0; green, 0; blue, 0 }  ][line width=0.75]    (10.93,-3.29) .. controls (6.95,-1.4) and (3.31,-0.3) .. (0,0) .. controls (3.31,0.3) and (6.95,1.4) .. (10.93,3.29)   ;
\draw  [fill={rgb, 255:red, 242; green, 237; blue, 171 }  ,fill opacity=1 ] (1010.5,687) -- (1070.5,724) -- (1010.5,761) -- (950.5,724) -- cycle ;
\draw  [fill={rgb, 255:red, 196; green, 246; blue, 142 }  ,fill opacity=1 ] (986.7,799) -- (1034.3,799) .. controls (1040.49,799) and (1045.5,807.95) .. (1045.5,819) .. controls (1045.5,830.05) and (1040.49,839) .. (1034.3,839) -- (986.7,839) .. controls (980.51,839) and (975.5,830.05) .. (975.5,819) .. controls (975.5,807.95) and (980.51,799) .. (986.7,799) -- cycle ;
\draw  [fill={rgb, 255:red, 236; green, 185; blue, 247 }  ,fill opacity=1 ] (774,688) -- (898.5,688) -- (898.5,763) -- (774,763) -- cycle ;
\draw    (302.5,651) -- (302.5,686) ;
\draw [shift={(302.5,688)}, rotate = 270] [color={rgb, 255:red, 0; green, 0; blue, 0 }  ][line width=0.75]    (10.93,-3.29) .. controls (6.95,-1.4) and (3.31,-0.3) .. (0,0) .. controls (3.31,0.3) and (6.95,1.4) .. (10.93,3.29)   ;
\draw  [fill={rgb, 255:red, 242; green, 237; blue, 171 }  ,fill opacity=1 ] (302.5,688) -- (362.5,725) -- (302.5,762) -- (242.5,725) -- cycle ;
\draw    (362.5,725) -- (419.5,725) ;
\draw [shift={(421.5,725)}, rotate = 180] [color={rgb, 255:red, 0; green, 0; blue, 0 }  ][line width=0.75]    (10.93,-3.29) .. controls (6.95,-1.4) and (3.31,-0.3) .. (0,0) .. controls (3.31,0.3) and (6.95,1.4) .. (10.93,3.29)   ;
\draw    (893.5,915) -- (942.5,915) ;
\draw [shift={(944.5,915)}, rotate = 180] [color={rgb, 255:red, 0; green, 0; blue, 0 }  ][line width=0.75]    (10.93,-3.29) .. controls (6.95,-1.4) and (3.31,-0.3) .. (0,0) .. controls (3.31,0.3) and (6.95,1.4) .. (10.93,3.29)   ;
\draw  [fill={rgb, 255:red, 196; green, 246; blue, 142 }  ,fill opacity=1 ] (955.7,895) -- (1003.3,895) .. controls (1009.49,895) and (1014.5,903.95) .. (1014.5,915) .. controls (1014.5,926.05) and (1009.49,935) .. (1003.3,935) -- (955.7,935) .. controls (949.51,935) and (944.5,926.05) .. (944.5,915) .. controls (944.5,903.95) and (949.51,895) .. (955.7,895) -- cycle ;
\draw    (834.5,949) -- (834.5,984) ;
\draw [shift={(834.5,986)}, rotate = 270] [color={rgb, 255:red, 0; green, 0; blue, 0 }  ][line width=0.75]    (10.93,-3.29) .. controls (6.95,-1.4) and (3.31,-0.3) .. (0,0) .. controls (3.31,0.3) and (6.95,1.4) .. (10.93,3.29)   ;
\draw  [fill={rgb, 255:red, 242; green, 237; blue, 171 }  ,fill opacity=1 ] (834.5,878) -- (894.5,915) -- (834.5,952) -- (774.5,915) -- cycle ;
\draw    (834.5,762) -- (834.5,876) ;
\draw [shift={(834.5,878)}, rotate = 270] [color={rgb, 255:red, 0; green, 0; blue, 0 }  ][line width=0.75]    (10.93,-3.29) .. controls (6.95,-1.4) and (3.31,-0.3) .. (0,0) .. controls (3.31,0.3) and (6.95,1.4) .. (10.93,3.29)   ;
\draw  [fill={rgb, 255:red, 196; green, 246; blue, 142 }  ,fill opacity=1 ] (277.2,797) -- (324.8,797) .. controls (330.99,797) and (336,805.95) .. (336,817) .. controls (336,828.05) and (330.99,837) .. (324.8,837) -- (277.2,837) .. controls (271.01,837) and (266,828.05) .. (266,817) .. controls (266,805.95) and (271.01,797) .. (277.2,797) -- cycle ;
\draw  [fill={rgb, 255:red, 242; green, 237; blue, 171 }  ,fill opacity=1 ] (481.5,688) -- (541.5,725) -- (481.5,762) -- (421.5,725) -- cycle ;
\draw    (481.5,762) -- (481.5,797) ;
\draw [shift={(481.5,799)}, rotate = 270] [color={rgb, 255:red, 0; green, 0; blue, 0 }  ][line width=0.75]    (10.93,-3.29) .. controls (6.95,-1.4) and (3.31,-0.3) .. (0,0) .. controls (3.31,0.3) and (6.95,1.4) .. (10.93,3.29)   ;
\draw    (542.5,725) -- (599.5,725) ;
\draw [shift={(601.5,725)}, rotate = 180] [color={rgb, 255:red, 0; green, 0; blue, 0 }  ][line width=0.75]    (10.93,-3.29) .. controls (6.95,-1.4) and (3.31,-0.3) .. (0,0) .. controls (3.31,0.3) and (6.95,1.4) .. (10.93,3.29)   ;
\draw  [fill={rgb, 255:red, 236; green, 185; blue, 247 }  ,fill opacity=1 ] (601,688) -- (725.5,688) -- (725.5,763) -- (601,763) -- cycle ;
\draw  [fill={rgb, 255:red, 242; green, 237; blue, 171 }  ,fill opacity=1 ] (661,802) -- (721,839) -- (661,876) -- (601,839) -- cycle ;
\draw    (661,765) -- (661,800) ;
\draw [shift={(661,802)}, rotate = 270] [color={rgb, 255:red, 0; green, 0; blue, 0 }  ][line width=0.75]    (10.93,-3.29) .. controls (6.95,-1.4) and (3.31,-0.3) .. (0,0) .. controls (3.31,0.3) and (6.95,1.4) .. (10.93,3.29)   ;
\draw  [fill={rgb, 255:red, 236; green, 185; blue, 247 }  ,fill opacity=1 ] (420,801) -- (544.5,801) -- (544.5,876) -- (420,876) -- cycle ;
\draw    (601.5,839) -- (548,838.04) ;
\draw [shift={(546,838)}, rotate = 1.03] [color={rgb, 255:red, 0; green, 0; blue, 0 }  ][line width=0.75]    (10.93,-3.29) .. controls (6.95,-1.4) and (3.31,-0.3) .. (0,0) .. controls (3.31,0.3) and (6.95,1.4) .. (10.93,3.29)   ;
\draw    (484,875) -- (484,979) ;
\draw [shift={(484,981)}, rotate = 270] [color={rgb, 255:red, 0; green, 0; blue, 0 }  ][line width=0.75]    (10.93,-3.29) .. controls (6.95,-1.4) and (3.31,-0.3) .. (0,0) .. controls (3.31,0.3) and (6.95,1.4) .. (10.93,3.29)   ;
\draw  [fill={rgb, 255:red, 242; green, 237; blue, 171 }  ,fill opacity=1 ] (484,982) -- (544,1019) -- (484,1056) -- (424,1019) -- cycle ;
\draw    (544,1019) -- (593,1019) ;
\draw [shift={(595,1019)}, rotate = 180] [color={rgb, 255:red, 0; green, 0; blue, 0 }  ][line width=0.75]    (10.93,-3.29) .. controls (6.95,-1.4) and (3.31,-0.3) .. (0,0) .. controls (3.31,0.3) and (6.95,1.4) .. (10.93,3.29)   ;
\draw  [fill={rgb, 255:red, 196; green, 246; blue, 142 }  ,fill opacity=1 ] (606.2,999) -- (653.8,999) .. controls (659.99,999) and (665,1007.95) .. (665,1019) .. controls (665,1030.05) and (659.99,1039) .. (653.8,1039) -- (606.2,1039) .. controls (600.01,1039) and (595,1030.05) .. (595,1019) .. controls (595,1007.95) and (600.01,999) .. (606.2,999) -- cycle ;
\draw    (483.5,1057) -- (483.5,1092) ;
\draw [shift={(483.5,1094)}, rotate = 270] [color={rgb, 255:red, 0; green, 0; blue, 0 }  ][line width=0.75]    (10.93,-3.29) .. controls (6.95,-1.4) and (3.31,-0.3) .. (0,0) .. controls (3.31,0.3) and (6.95,1.4) .. (10.93,3.29)   ;
\draw  [fill={rgb, 255:red, 196; green, 246; blue, 142 }  ,fill opacity=1 ] (459.2,1095) -- (506.8,1095) .. controls (512.99,1095) and (518,1103.95) .. (518,1115) .. controls (518,1126.05) and (512.99,1135) .. (506.8,1135) -- (459.2,1135) .. controls (453.01,1135) and (448,1126.05) .. (448,1115) .. controls (448,1103.95) and (453.01,1095) .. (459.2,1095) -- cycle ;
\draw  [fill={rgb, 255:red, 196; green, 246; blue, 142 }  ,fill opacity=1 ] (810.2,986) -- (857.8,986) .. controls (863.99,986) and (869,994.95) .. (869,1006) .. controls (869,1017.05) and (863.99,1026) .. (857.8,1026) -- (810.2,1026) .. controls (804.01,1026) and (799,1017.05) .. (799,1006) .. controls (799,994.95) and (804.01,986) .. (810.2,986) -- cycle ;
\draw    (1010.5,761) -- (1010.5,796) ;
\draw [shift={(1010.5,798)}, rotate = 270] [color={rgb, 255:red, 0; green, 0; blue, 0 }  ][line width=0.75]    (10.93,-3.29) .. controls (6.95,-1.4) and (3.31,-0.3) .. (0,0) .. controls (3.31,0.3) and (6.95,1.4) .. (10.93,3.29)   ;
\draw    (302.5,760) -- (302.5,795) ;
\draw [shift={(302.5,797)}, rotate = 270] [color={rgb, 255:red, 0; green, 0; blue, 0 }  ][line width=0.75]    (10.93,-3.29) .. controls (6.95,-1.4) and (3.31,-0.3) .. (0,0) .. controls (3.31,0.3) and (6.95,1.4) .. (10.93,3.29)   ;
\draw  [fill={rgb, 255:red, 196; green, 246; blue, 142 }  ,fill opacity=1 ] (636.7,911) -- (684.3,911) .. controls (690.49,911) and (695.5,919.95) .. (695.5,931) .. controls (695.5,942.05) and (690.49,951) .. (684.3,951) -- (636.7,951) .. controls (630.51,951) and (625.5,942.05) .. (625.5,931) .. controls (625.5,919.95) and (630.51,911) .. (636.7,911) -- cycle ;
\draw    (660.5,873) -- (660.5,908) ;
\draw [shift={(660.5,910)}, rotate = 270] [color={rgb, 255:red, 0; green, 0; blue, 0 }  ][line width=0.75]    (10.93,-3.29) .. controls (6.95,-1.4) and (3.31,-0.3) .. (0,0) .. controls (3.31,0.3) and (6.95,1.4) .. (10.93,3.29)   ;
\draw  [fill={rgb, 255:red, 242; green, 237; blue, 171 }  ,fill opacity=1 ] (495.5,246) -- (555.5,283) -- (495.5,320) -- (435.5,283) -- cycle ;
\draw    (553.5,283) -- (602.5,283) ;
\draw [shift={(604.5,283)}, rotate = 180] [color={rgb, 255:red, 0; green, 0; blue, 0 }  ][line width=0.75]    (10.93,-3.29) .. controls (6.95,-1.4) and (3.31,-0.3) .. (0,0) .. controls (3.31,0.3) and (6.95,1.4) .. (10.93,3.29)   ;
\draw    (436.5,283) -- (383,282.04) ;
\draw [shift={(381,282)}, rotate = 1.03] [color={rgb, 255:red, 0; green, 0; blue, 0 }  ][line width=0.75]    (10.93,-3.29) .. controls (6.95,-1.4) and (3.31,-0.3) .. (0,0) .. controls (3.31,0.3) and (6.95,1.4) .. (10.93,3.29)   ;
\draw  [fill={rgb, 255:red, 196; green, 246; blue, 142 }  ,fill opacity=1 ] (319.7,262) -- (367.3,262) .. controls (373.49,262) and (378.5,270.95) .. (378.5,282) .. controls (378.5,293.05) and (373.49,302) .. (367.3,302) -- (319.7,302) .. controls (313.51,302) and (308.5,293.05) .. (308.5,282) .. controls (308.5,270.95) and (313.51,262) .. (319.7,262) -- cycle ;

\draw (440,155) node [anchor=north west][inner sep=0.75pt]   [align=left] {\begin{minipage}[lt]{72.55pt}\setlength\topsep{0pt}
\begin{center}
{\scriptsize Parse dispositive }\\{\scriptsize variables from all CCs}
\end{center}

\end{minipage}};
\draw (643,268) node [anchor=north west][inner sep=0.75pt]   [align=left] {\begin{minipage}[lt]{29.71pt}\setlength\topsep{0pt}
\begin{center}
{\scriptsize In scope}\\{\scriptsize of Act?}
\end{center}

\end{minipage}};
\draw (777,266) node [anchor=north west][inner sep=0.75pt]   [align=left] {\begin{minipage}[lt]{41.62pt}\setlength\topsep{0pt}
\begin{center}
{\scriptsize Out of }\\{\scriptsize scope of Act}
\end{center}

\end{minipage}};
\draw (734,263) node [anchor=north west][inner sep=0.75pt]   [align=left] {\begin{minipage}[lt]{11.85pt}\setlength\topsep{0pt}
\begin{center}
{\scriptsize No}
\end{center}

\end{minipage}};
\draw (635,323) node [anchor=north west][inner sep=0.75pt]   [align=left] {\begin{minipage}[lt]{14.37pt}\setlength\topsep{0pt}
\begin{center}
{\scriptsize Yes}
\end{center}

\end{minipage}};
\draw (634,388) node [anchor=north west][inner sep=0.75pt]   [align=left] {\begin{minipage}[lt]{38.84pt}\setlength\topsep{0pt}
\begin{center}
{\scriptsize Exempted?}
\end{center}

\end{minipage}};
\draw (780,377) node [anchor=north west][inner sep=0.75pt]   [align=left] {\begin{minipage}[lt]{41.62pt}\setlength\topsep{0pt}
\begin{center}
{\scriptsize Out of }\\{\scriptsize scope of Act}
\end{center}

\end{minipage}};
\draw (737,374) node [anchor=north west][inner sep=0.75pt]   [align=left] {\begin{minipage}[lt]{14.37pt}\setlength\topsep{0pt}
\begin{center}
{\scriptsize Yes}
\end{center}

\end{minipage}};
\draw (635,435) node [anchor=north west][inner sep=0.75pt]   [align=left] {\begin{minipage}[lt]{11.85pt}\setlength\topsep{0pt}
\begin{center}
{\scriptsize No}
\end{center}

\end{minipage}};
\draw (639,490) node [anchor=north west][inner sep=0.75pt]   [align=left] {\begin{minipage}[lt]{36.86pt}\setlength\topsep{0pt}
\begin{center}
{\scriptsize Prohibited }\\{\scriptsize practice?}
\end{center}

\end{minipage}};
\draw (785.7,492) node [anchor=north west][inner sep=0.75pt]   [align=left] {\begin{minipage}[lt]{33.28pt}\setlength\topsep{0pt}
\begin{center}
{\scriptsize Not }\\{\scriptsize compliant}
\end{center}

\end{minipage}};
\draw (735,486) node [anchor=north west][inner sep=0.75pt]   [align=left] {\begin{minipage}[lt]{14.37pt}\setlength\topsep{0pt}
\begin{center}
{\scriptsize Yes}
\end{center}

\end{minipage}};
\draw (635,543) node [anchor=north west][inner sep=0.75pt]   [align=left] {\begin{minipage}[lt]{11.85pt}\setlength\topsep{0pt}
\begin{center}
{\scriptsize No}
\end{center}

\end{minipage}};
\draw (642,599) node [anchor=north west][inner sep=0.75pt]   [align=left] {\begin{minipage}[lt]{28.12pt}\setlength\topsep{0pt}
\begin{center}
{\scriptsize Type of }\\{\scriptsize project?}
\end{center}

\end{minipage}};
\draw (723,580) node [anchor=north west][inner sep=0.75pt]   [align=left] {\begin{minipage}[lt]{22.17pt}\setlength\topsep{0pt}
\begin{center}
{\scriptsize GPAI }\\{\scriptsize model}
\end{center}

\end{minipage}};
\draw (560,579) node [anchor=north west][inner sep=0.75pt]   [align=left] {\begin{minipage}[lt]{25.34pt}\setlength\topsep{0pt}
\begin{center}
{\scriptsize AI }\\{\scriptsize system}
\end{center}

\end{minipage}};
\draw (809,599) node [anchor=north west][inner sep=0.75pt]   [align=left] {\begin{minipage}[lt]{33.67pt}\setlength\topsep{0pt}
\begin{center}
{\scriptsize Systemic }\\{\scriptsize Risk?}
\end{center}

\end{minipage}};
\draw (457,607) node [anchor=north west][inner sep=0.75pt]   [align=left] {\begin{minipage}[lt]{34.86pt}\setlength\topsep{0pt}
\begin{center}
{\scriptsize High-risk?}
\end{center}

\end{minipage}};
\draw (905,594) node [anchor=north west][inner sep=0.75pt]   [align=left] {\begin{minipage}[lt]{14.37pt}\setlength\topsep{0pt}
\begin{center}
{\scriptsize Yes}
\end{center}

\end{minipage}};
\draw (803,654) node [anchor=north west][inner sep=0.75pt]   [align=left] {\begin{minipage}[lt]{11.85pt}\setlength\topsep{0pt}
\begin{center}
{\scriptsize No}
\end{center}

\end{minipage}};
\draw (945,582) node [anchor=north west][inner sep=0.75pt]   [align=left] {\begin{minipage}[lt]{85.27pt}\setlength\topsep{0pt}
\begin{center}
{\scriptsize Check all CCs to see that }\\{\scriptsize requirements for GPAI }\\{\scriptsize models with }\\{\scriptsize systemic risk are met }
\end{center}

\end{minipage}};
\draw (456,49.8) node [anchor=north west][inner sep=0.75pt]   [align=left] {\begin{minipage}[lt]{35.66pt}\setlength\topsep{0pt}
\begin{center}
{\scriptsize Full set of }\\{\scriptsize CCs}
\end{center}

\end{minipage}};
\draw (390,593) node [anchor=north west][inner sep=0.75pt]   [align=left] {\begin{minipage}[lt]{14.37pt}\setlength\topsep{0pt}
\begin{center}
{\scriptsize Yes}
\end{center}

\end{minipage}};
\draw (452,654) node [anchor=north west][inner sep=0.75pt]   [align=left] {\begin{minipage}[lt]{11.85pt}\setlength\topsep{0pt}
\begin{center}
{\scriptsize No}
\end{center}

\end{minipage}};
\draw (243,584) node [anchor=north west][inner sep=0.75pt]   [align=left] {\begin{minipage}[lt]{85.27pt}\setlength\topsep{0pt}
\begin{center}
{\scriptsize Check all CCs to see that }\\{\scriptsize requirements for }\\{\scriptsize high-risk AI \ }\\{\scriptsize systemic risk are met }
\end{center}

\end{minipage}};
\draw (976,709) node [anchor=north west][inner sep=0.75pt]   [align=left] {\begin{minipage}[lt]{49.15pt}\setlength\topsep{0pt}
\begin{center}
{\scriptsize Requirements }\\{\scriptsize met?}
\end{center}

\end{minipage}};
\draw (916,703) node [anchor=north west][inner sep=0.75pt]   [align=left] {\begin{minipage}[lt]{14.37pt}\setlength\topsep{0pt}
\begin{center}
{\scriptsize Yes}
\end{center}

\end{minipage}};
\draw (985.7,804) node [anchor=north west][inner sep=0.75pt]   [align=left] {\begin{minipage}[lt]{33.28pt}\setlength\topsep{0pt}
\begin{center}
{\scriptsize Not }\\{\scriptsize compliant}
\end{center}

\end{minipage}};
\draw (776,705) node [anchor=north west][inner sep=0.75pt]   [align=left] {\begin{minipage}[lt]{79.71pt}\setlength\topsep{0pt}
\begin{center}
{\scriptsize Check all CCs to see }\\{\scriptsize that requirements for all }\\{\scriptsize GPAI models are met }
\end{center}

\end{minipage}};
\draw (268,711) node [anchor=north west][inner sep=0.75pt]   [align=left] {\begin{minipage}[lt]{49.15pt}\setlength\topsep{0pt}
\begin{center}
{\scriptsize Requirements }\\{\scriptsize met?}
\end{center}

\end{minipage}};
\draw (374,705) node [anchor=north west][inner sep=0.75pt]   [align=left] {\begin{minipage}[lt]{14.37pt}\setlength\topsep{0pt}
\begin{center}
{\scriptsize Yes}
\end{center}

\end{minipage}};
\draw (954.7,900) node [anchor=north west][inner sep=0.75pt]   [align=left] {\begin{minipage}[lt]{33.28pt}\setlength\topsep{0pt}
\begin{center}
{\scriptsize Not }\\{\scriptsize compliant}
\end{center}

\end{minipage}};
\draw (905,895) node [anchor=north west][inner sep=0.75pt]   [align=left] {\begin{minipage}[lt]{11.85pt}\setlength\topsep{0pt}
\begin{center}
{\scriptsize No}
\end{center}

\end{minipage}};
\draw (805,952) node [anchor=north west][inner sep=0.75pt]   [align=left] {\begin{minipage}[lt]{14.37pt}\setlength\topsep{0pt}
\begin{center}
{\scriptsize Yes}
\end{center}

\end{minipage}};
\draw (800,903) node [anchor=north west][inner sep=0.75pt]   [align=left] {\begin{minipage}[lt]{49.15pt}\setlength\topsep{0pt}
\begin{center}
{\scriptsize Requirements }\\{\scriptsize met?}
\end{center}

\end{minipage}};
\draw (276.2,802) node [anchor=north west][inner sep=0.75pt]   [align=left] {\begin{minipage}[lt]{33.28pt}\setlength\topsep{0pt}
\begin{center}
{\scriptsize Not }\\{\scriptsize compliant}
\end{center}

\end{minipage}};
\draw (451,710) node [anchor=north west][inner sep=0.75pt]   [align=left] {\begin{minipage}[lt]{41.61pt}\setlength\topsep{0pt}
\begin{center}
{\scriptsize Is Article 50 }\\{\scriptsize invoked?}
\end{center}

\end{minipage}};
\draw (452,765) node [anchor=north west][inner sep=0.75pt]   [align=left] {\begin{minipage}[lt]{11.85pt}\setlength\topsep{0pt}
\begin{center}
{\scriptsize No}
\end{center}

\end{minipage}};
\draw (554,705) node [anchor=north west][inner sep=0.75pt]   [align=left] {\begin{minipage}[lt]{14.37pt}\setlength\topsep{0pt}
\begin{center}
{\scriptsize Yes}
\end{center}

\end{minipage}};
\draw (605,696) node [anchor=north west][inner sep=0.75pt]   [align=left] {\begin{minipage}[lt]{76.55pt}\setlength\topsep{0pt}
\begin{center}
{\scriptsize Check Project CC to }\\{\scriptsize see that Article 50}\\{\scriptsize additional transparency}\\{\scriptsize requirements are met }
\end{center}

\end{minipage}};
\draw (622.5,826) node [anchor=north west][inner sep=0.75pt]   [align=left] {\begin{minipage}[lt]{49.15pt}\setlength\topsep{0pt}
\begin{center}
{\scriptsize Requirements }\\{\scriptsize met?}
\end{center}

\end{minipage}};
\draw (427,818) node [anchor=north west][inner sep=0.75pt]   [align=left] {\begin{minipage}[lt]{73.76pt}\setlength\topsep{0pt}
\begin{center}
{\scriptsize Check Project CC to }\\{\scriptsize see that requirements }\\{\scriptsize for all AI systems met}
\end{center}

\end{minipage}};
\draw (566,818) node [anchor=north west][inner sep=0.75pt]   [align=left] {\begin{minipage}[lt]{14.37pt}\setlength\topsep{0pt}
\begin{center}
{\scriptsize Yes}
\end{center}

\end{minipage}};
\draw (449.5,1006) node [anchor=north west][inner sep=0.75pt]   [align=left] {\begin{minipage}[lt]{49.15pt}\setlength\topsep{0pt}
\begin{center}
{\scriptsize Requirements }\\{\scriptsize met?}
\end{center}

\end{minipage}};
\draw (606.2,1005) node [anchor=north west][inner sep=0.75pt]   [align=left] {\begin{minipage}[lt]{33.28pt}\setlength\topsep{0pt}
\begin{center}
{\scriptsize Not }\\{\scriptsize compliant}
\end{center}

\end{minipage}};
\draw (555.5,999) node [anchor=north west][inner sep=0.75pt]   [align=left] {\begin{minipage}[lt]{11.85pt}\setlength\topsep{0pt}
\begin{center}
{\scriptsize No}
\end{center}

\end{minipage}};
\draw (454,1060) node [anchor=north west][inner sep=0.75pt]   [align=left] {\begin{minipage}[lt]{14.37pt}\setlength\topsep{0pt}
\begin{center}
{\scriptsize Yes}
\end{center}

\end{minipage}};
\draw (457.2,1105) node [anchor=north west][inner sep=0.75pt]   [align=left] {\begin{minipage}[lt]{34.87pt}\setlength\topsep{0pt}
\begin{center}
{\scriptsize Compliant}
\end{center}

\end{minipage}};
\draw (808.2,996) node [anchor=north west][inner sep=0.75pt]   [align=left] {\begin{minipage}[lt]{34.87pt}\setlength\topsep{0pt}
\begin{center}
{\scriptsize Compliant}
\end{center}

\end{minipage}};
\draw (273,763) node [anchor=north west][inner sep=0.75pt]   [align=left] {\begin{minipage}[lt]{11.85pt}\setlength\topsep{0pt}
\begin{center}
{\scriptsize No}
\end{center}

\end{minipage}};
\draw (635.7,916) node [anchor=north west][inner sep=0.75pt]   [align=left] {\begin{minipage}[lt]{33.28pt}\setlength\topsep{0pt}
\begin{center}
{\scriptsize Not }\\{\scriptsize compliant}
\end{center}

\end{minipage}};
\draw (460,265) node [anchor=north west][inner sep=0.75pt]   [align=left] {\begin{minipage}[lt]{49.54pt}\setlength\topsep{0pt}
\begin{center}
{\scriptsize In scope}\\{\scriptsize of CC system?}
\end{center}

\end{minipage}};
\draw (564,259) node [anchor=north west][inner sep=0.75pt]   [align=left] {\begin{minipage}[lt]{14.37pt}\setlength\topsep{0pt}
\begin{center}
{\scriptsize Yes}
\end{center}

\end{minipage}};
\draw (401,262) node [anchor=north west][inner sep=0.75pt]   [align=left] {\begin{minipage}[lt]{11.85pt}\setlength\topsep{0pt}
\begin{center}
{\scriptsize No}
\end{center}

\end{minipage}};
\draw (312,265) node [anchor=north west][inner sep=0.75pt]   [align=left] {\begin{minipage}[lt]{42.01pt}\setlength\topsep{0pt}
\begin{center}
{\scriptsize Out of }\\{\scriptsize scope of CC}
\end{center}

\end{minipage}};
\draw (635,879) node [anchor=north west][inner sep=0.75pt]   [align=left] {\begin{minipage}[lt]{11.85pt}\setlength\topsep{0pt}
\begin{center}
{\scriptsize No}
\end{center}

\end{minipage}};
\draw (986,767) node [anchor=north west][inner sep=0.75pt]   [align=left] {\begin{minipage}[lt]{11.85pt}\setlength\topsep{0pt}
\begin{center}
{\scriptsize No}
\end{center}

\end{minipage}};

\end{tikzpicture}

}

\end{center}

\end{document}